\newcommand{\CLASSINPUTtoptextmargin}{21.0mm}
\newcommand{\CLASSINPUTbottomtextmargin}{15.5mm}
\newcommand{\CLASSINPUTinnersidemargin}{17.5mm}
\newcommand{\CLASSINPUToutersidemargin}{17.5mm}

\documentclass[letterpaper, 10 pt, journal, twoside]{IEEEtran}  

\makeatletter
\let\NAT@parse\undefined
\makeatother
\newcommand{\myTitle}{CNN based Road User Detection using the 3D Radar Cube} 
\newcommand{\myKeywords}{CNN, Radar, Road Users, Radar Cube} 
\usepackage[pdftex,
			pdfauthor={Andras Palffy},
			pdftitle={\myTitle},
			pdfkeywords={\myKeywords},
			]{hyperref} 
\usepackage{graphics} 
\usepackage{epsfig} 
\usepackage{times} 
\usepackage{amsmath} 
\usepackage{amssymb}  
\usepackage{todonotes} 
\usepackage{textcomp} 
\usepackage{float}
\usepackage[subrefformat=parens, labelformat=parens]{subcaption} 
\usepackage{booktabs}
\usepackage{gensymb}
\usepackage{tikz}
\usetikzlibrary{arrows,shapes,automata,petri,positioning,calc}
\usepackage{pgfplots} 
\usepackage{xspace}
\usepackage{adjustbox}
\usepackage{multirow}
\usepackage{soul}
\pgfplotsset{compat=1.3}

\usepackage{color}      
\usepackage{soul} 

\usepackage[sort,nocompress]{cite} 
\hypersetup{
	colorlinks,
	linkcolor=black,
	citecolor=black,
	urlcolor=black
}

\newcommand{\rangehigh}{r}
\newcommand{\azimuthhigh}{\alpha}

\newcommand{\velr}{v_r}
\newcommand{\RCS}{RCS}

\newcommand{\ChannelNum}{C}

\newcommand{\dopplercroprange}{H}
\newcommand{\rangecroprange}{L}
\newcommand{\azimuthcroprange}{W}
\newcommand{\MinPoints}{MinPoints}
\newcommand{\MinSpeed}{v_{min}}
\newcommand{\spatialRadius}{\gamma_{xy}}
\newcommand{\velocityRadius}{\gamma_{v}}

\newcommand{\LLTensemble}{\textit{RTCnet}\xspace}
\newcommand{\LLTmulticlass}{\textit{RTCnet (no ensemble)}\xspace}
\newcommand{\LLTnospeed}{\textit{RTCnet (no speed)}\xspace}
\newcommand{\LLTnoRCS}{\textit{RTCnet (no RCS)}\xspace}
\newcommand{\LLThigh}{\textit{RTCnet (no low-level)}\xspace}
\newcommand{\baselineSchumann}{\textit{Schumann} \cite{Schumann2017}\xspace}
\newcommand{\baselineProphet}{\textit{Prophet} \cite{Prophet}\xspace}

\newcommand{\rev}[1]{{{#1}}} 

\newcommand{\DMG}[1]{{{#1}}} 

\title{
	\myTitle
}

\author{Andras Palffy\,$^{1}$, Jiaao Dong\,$^{1}$, Julian F. P. Kooij\,$^{1}$ and Dariu M. Gavrila\,$^{1}$
\thanks{Manuscript received: September 10, 2019; Revised December 12, 2019; Accepted January 9, 2020.}
\thanks{This paper was recommended for publication by Editor Eric Marchand upon evaluation of the Associate Editor and Reviewers' comments.} 
\thanks{$^{1}$All authors are with Intelligent Vehicles Group, Delft University of Technology, The Netherlands
	{\tt\footnotesize a.palffy@tudelft.nl}}%
\thanks{Digital Object Identifier (DOI): see top of this page.}
}
\begin{document}

\maketitle

\markboth{IEEE Robotics and Automation Letters. Preprint Version. Accepted January, 2020}
{Palffy \MakeLowercase{\textit{et al.}}: \myTitle}


\begin{abstract}
This paper presents a novel radar based, single-frame, multi-class detection method for {moving} road users (\textit{pedestrian, cyclist, car}), which utilizes low-level {radar cube data}.
The method {provides} class information both on the {radar target- and object-level}. {Radar targets are classified} individually 
after extending the target features with a cropped block of the 3D radar cube around their positions, 
thereby capturing the motion of moving parts in the local velocity distribution.
A Convolutional Neural Network (CNN) is proposed for this classification step.
Afterwards, object proposals are {generated} with a 
clustering step, which not only considers the radar targets' positions and velocities, but their calculated class scores as well.
In experiments on a real-life dataset we demonstrate that our method outperforms the state-of-the-art methods both target- and object-wise by reaching an average of \rev{0.70} (baseline: \rev{0.68}) target-wise and \rev{0.56} (baseline: \rev{0.48}) object-wise F1 score.
Furthermore, we examine the importance of the used features in {an ablation} study.  \looseness=-1
\end{abstract}
\section{INTRODUCTION}

\IEEEPARstart{R}{adars} are attractive sensors for intelligent vehicles as they are {relatively} robust to weather and lighting conditions (e.g. rain, snow, darkness) compared to camera and LIDAR sensors. Radars also have excellent range sensitivity and can measure {radial object velocities} directly using the Doppler effect. 
{Thus, they are widely used in applications such as adaptive cruise control and pre-crash safety}. 


Commercially available radars output a point-cloud of reflections called \textit{radar targets} in every frame (sweep). Each radar target has the following features: range $\rangehigh $ and azimuth $ \azimuthhigh $, {radar cross section} $ \RCS $ {(i.e. reflectivity)}, and the object's radial speed $ \velr $ relative to the ego-vehicle. We will call these features \textit{target-level}.
%
Since a single reflection {does not convey} enough information to {segment and} classify {an entire object}, many radar based road user detection methods (e.g. \cite{Prophet,Schumann2017,Scheiner2018}) first cluster radar targets by their {target-level features}. 
Clusters are then classified as a whole based \rev{on} derived statistical features (e.g. mean, variance of $ \rangehigh, \velr, \RCS $ of contained radar targets), and the same class label is assigned to all radar targets in the cluster. {Object segmentation and classification performance in such pipeline \DMG{depend} on the success of the initial clustering step}.

\begin{figure}
	\centering
	\vspace{0pt}
	\includegraphics[width = 0.9\linewidth]{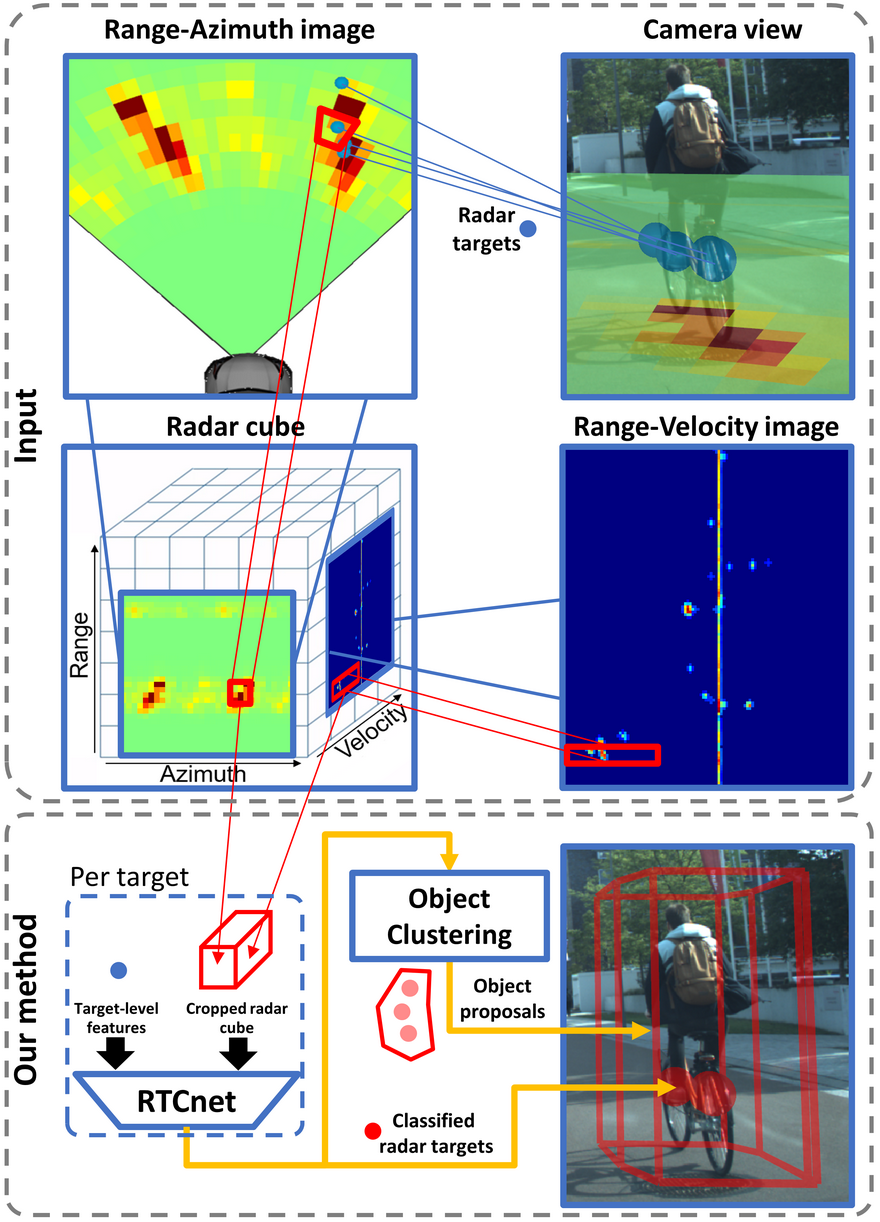}
	\caption{
	Inputs (radar cube and radar targets, top), main processing blocks (RTCnet and object clustering, bottom left), and outputs (classified radar targets and object proposals, bottom right) of our proposed method. Classified radar targets are shown as colored spheres at the sensor's height. Object proposals are visualized by a convex hull around the clustered targets on the ground plane and at $ 2~m $.
	}
	\label{fig:opening}
\end{figure}

\begin{figure}[t]
	\centering
	\includegraphics[width = 0.8\linewidth]{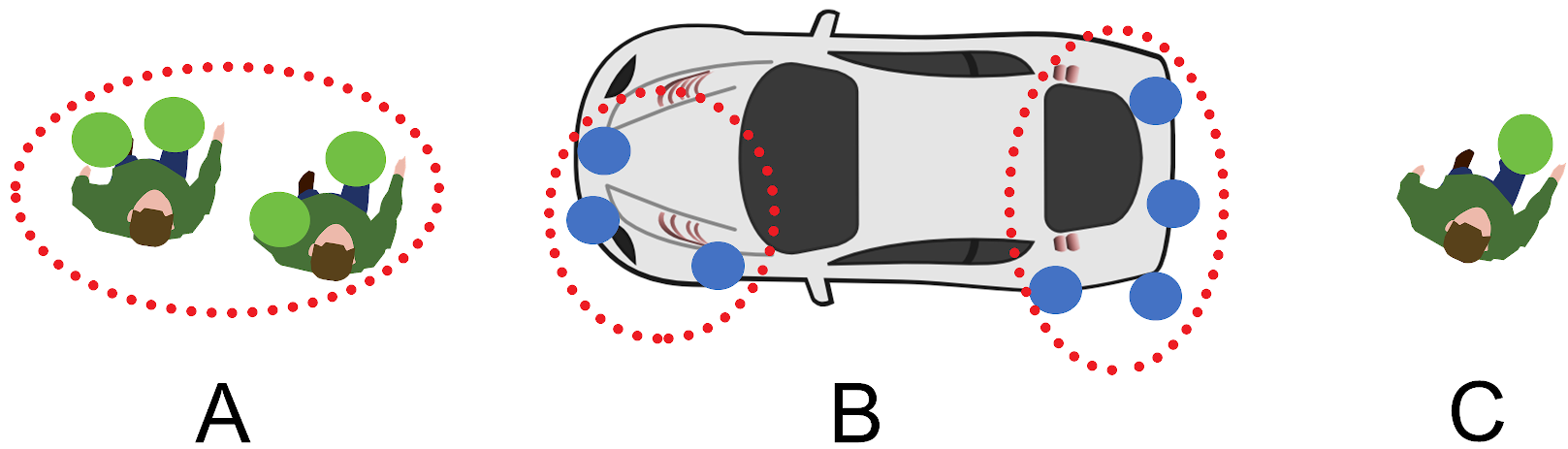}
	\caption{
		Challenging cases for cluster-wise classification methods. 
		A: Objects may be clustered together (red circle).
		B: Large objects may be split up into several clusters.
		C: Object with only one reflection. 
		Radar targets are shown as dots, 
		colored green/blue for pedestrian/car ground truth class.}
	\label{fig:Clustering_errors}
\end{figure}

Various methods \cite{Schubert2014,Schubert2015, Angelov2018} instead explore using the \textit{low-level} \textit{radar cube}
extracted from an earlier signal processing stage of the radar.
The \DMG{radar} cube is a 3D data matrix with axes corresponding to range, azimuth, and velocity (also called Doppler),
and a cell's value represents the measured radar reflectivity in that range/azimuth/Doppler bin.
In contrast to the target-level data,
the radar cube provides the complete speed distribution (i.e.~Doppler vector) at multiple 2D range-azimuth locations. Such distributions can capture modulations of an object's main velocity caused by its moving parts, e.g. swinging limbs or rotating wheels, and were shown to be a valuable feature for object classification \cite{Schubert2014,Schubert2015}.
Commonly radar cube features are computed by first
generating 2D range-azimuth or range-Doppler projections,
or by aggregating the projected Doppler axis over time into a Doppler-time image \cite{Angelov2018, Kwon2017}.
We will call features derived from the 3D cube or its projections \textit{low-level}.
A downside of such low-level radar data is the lower range and azimuth resolution than the radar targets, and that radar phase ambiguity is not yet addressed, since no advanced range interpolation and direction-of-arrival estimation has taken place. 

In this paper we propose a radar based, multi-class moving road user detection method, which exploits \textit{both} expert knowledge at the target-level (accurate 2D location, addressed phase ambiguity), and low-level information from the \textit{full} 3D radar cube rather than a 2D projection.
Importantly, the inclusion of low-level data enables classification of individual radar targets before any \DMG{object clustering; the latter step} can benefit from the obtained class scores.
At the core of our method is a {Convolutional Neural Network (CNN)} called Radar {Target} Classification Network, or \LLTensemble for short.
See Fig. \ref{fig:opening} for an overview of our method's inputs (radar targets and cube) and outputs (classified targets and object proposals). 

Our method can provide class information on both radar target-level and object-level.
Target-level class labels are valuable for sensor fusion operating on intermediate-level, i.e. handling multiple measurements per object \cite{Palffy2019,Granstrom2017}.
Our {target-level} classification
is more robust than cluster-wise classification where
the initial clustering step must manage to separate radar targets from different objects, and keep those coming from the same object together, see Fig. \ref{fig:Clustering_errors}.
Our object-level class information provides instances that are both segmented and classified (object detection), which is valuable for high-level (i.e. late) sensor fusion.
While traditional methods must perform clustering with a single set of parameters for all classes,
our approach enables use of class-specific clustering parameters (e.g. larger {object} radius for cars).


\section{RELATED WORK}

\begin{table}[h]
	\centering
	\begin{tabular}{@{}lllll@{}} \toprule
		Method                          & Basis     & Features & Classes & Time window       \\ \midrule
		\baselineProphet   $^\dagger$ & clusters & target     & single  & 1 frame (50 ms)   \\ 
		\baselineSchumann $^\dagger$  & clusters & target     & multi   & 2 frames (150 ms) \\ 
		Prophet \cite{Prophet2018}    & clusters & both     & single  & 1 frame           \\ 
		Schumann \cite{Schumann2018}  & targets  & target     & multi   & 0.5 s             \\ 
		Angelov \cite{Angelov2018}    & targets  & low      & multi   & 0.5-2 s           \\ 
		\LLTensemble \textit{(ours)}           & \textit{targets}  & \textit{both}     & \textit{multi}   & \textit{1 frame (75 ms)} \\ \bottomrule
	\end{tabular}
	\caption{{Overview of the most closely-related methods.} \\
	{$^\dagger$: marks} methods selected as baselines.
 }
	\label{tab:overview_of_methods}
\end{table}

%
{Some previous work on radar in automotive setting has dealt with \textit{static} environments.} E.g. \cite{Brodeski2019DeepRD} shows preliminary results of a neural network based method in a static experimental setup, which creates accurate target-level information from the radar cube. \cite{ICRASLAM2019} creates an occupancy grid with low-level data. {Static object classification} (e.g. parked cars, traffic signs) has been shown with target-level \cite{Lombacher2016} and with low-level data \cite{Patel}. 
We will focus only on methods addressing \textit{moving} road users.

Many road user detection methods start by clustering the radar targets into a set of object proposals. 
In \cite{Prophet}, radar targets are first clustered into objects by DBSCAN\cite{Ester2010}. Then, several cluster-wise features are extracted, e.g. the variance/mean of $ \velr$ and $ \rangehigh$. {The performance of various classifiers (Random Forest, Support Vector Machine (SVM), 1-layer Neural Network, etc.) were compared} in a single-class (pedestrian) detection task.
\cite{Schumann2017} also uses clusters calculated by DBSCAN as the base of a multi-class (\textit{car, pedestrian, group of pedestrians, cyclist, truck}) detection, but extract different features, e.g. deviation and spread of $ \azimuthhigh$. 
Afterwards, Long Short-Term Memory (LSTM) and Random Forest classifiers were compared for the classification step. 
Falsely merged clusters (Fig. \ref{fig:Clustering_errors}, A) were corrected manually to focus on the classification task itself. 
The same authors showed a method \cite{Schumann2018a} to incorporate a priori knowledge about the data into the clustering. \cite{Scheiner2019a} also aims to improve the clustering with a multi-stage approach.
\cite{Scheiner2018} follows the work of \cite{Schumann2017} for clustering and classification, {but tests and ranks further cluster-wise features in a backward elimination study.

While clustering based methods are widely used, it is often noted (e.g. \cite{Schumann2018}, \cite{Schumann2018a}) that the clustering step {is error-prone}. Objects can be mistakenly merged (Fig. \ref{fig:Clustering_errors}, A) or split apart (Fig. \ref{fig:Clustering_errors}, B). 
Finding {suitable} parameters (e.g. radius and minimum number of points for DBSCAN) is challenging as the same parameters must be used for all classes, although they have significantly different spatial extension and velocity profiles. E.g. a larger radius is beneficial for cars, but could falsely merge pedestrians and cyclists.
Another challenge of clustering based methods is that small objects may not have enough reflections (Fig. \ref{fig:Clustering_errors}, C) to extract meaningful statistical features, e.g. variance. E.g. both \cite{Prophet} and \cite{Schumann2017} have DBSCAN's minimum number of points to form a cluster ($ \MinPoints $) larger than one, which means that single standing points are thrown away. \looseness=-1

To address these challenges, there is a trend to classify each target individually instead of in clusters.
Encouraged by the results achieved with semantic segmentation networks on point-clouds from LIDAR or stereo camera setups, e.g. Pointnet++ \cite{Qi2017b}, researchers have tried to apply the same techniques to radar data.
However, the output of a single radar sweep is too sparse. To overcome this, they used multiple frames \cite{Schumann2018} or addressed large road users (cars) only \cite{Danzer2019}.
%

Low-level radar data has been used for road user classification, especially for pedestrians. 
E.g. a walking pedestrian's Doppler-time image contains a characteristic walking gait pattern \cite{Schubert2014,Schubert2015}. 
This is beneficial to exploit if the radar sensor is stationary, e.g. in surveillance applications \cite{DaveTahmoushandJerrySilvious}, \cite{Okumura2016}, \cite{Kwon2017}. 
%
Doppler-time features were also used in automotive setups.
\cite{Angelov2018} applies a CNN-LSTM network on Range-Doppler and Doppler-Time spectrograms of 0.5-2 seconds to classify \textit{pedestrian, group of pedestrians, car}, and \textit{cyclist} classes. 
%
\cite{Prophet2018} pointed out that a long multi-frame observation period is not viable for urban driving, and proposed a single-frame usage of low-level data.
Their method still generates object proposals with DBSCAN similar to  \cite{Prophet,Schumann2017},
but extracts for each cluster the corresponding area in a 2D Range-Doppler image, which is then classified using conventional computer vision. 
%
In \cite{Perez2018}, the full radar cube is used as a multi-channel image input to a CNN network to classify \textit{cars, pedestrians}, and \textit{cyclists}. The study only addresses a single-object classification task, i.e. location is not fetched.

In conclusion, the topic of radar based road user detection was extensively researched. 
Table \ref{tab:overview_of_methods} gives an overview of the most relevant methods with their basis of the classification (cluster-wise or target-wise), the level of features (target or low), the number of classified classes, and the required time window to collect suitable amount of data.
None of the found methods {avoids error-prone clustering for classification} and operates with a {low} latency for urban driving (i.e. one or two radar sweeps ($75-150~ms$)) at the same time. 
%

Our main contributions are as follows. 
1) We propose a radar based, {single-frame, multi-class} (\textit{pedestrian, cyclist, car}) moving road user detection method, which exploits both target-level and low-level radar data by a specially designed CNN. The method provides both classified radar targets and object proposals by a class-specific clustering.
2) We show on a large-scale, real-world dataset that our method is able to detect road users with higher than state-of-the-art performance both in target-wise (target classification) and object-wise (object detection) metrics using only a single frame of radar data.
\begin{figure*}
	\centering
	\includegraphics[width = 0.94\linewidth, trim={0 0pt 0 0pt}, clip,]{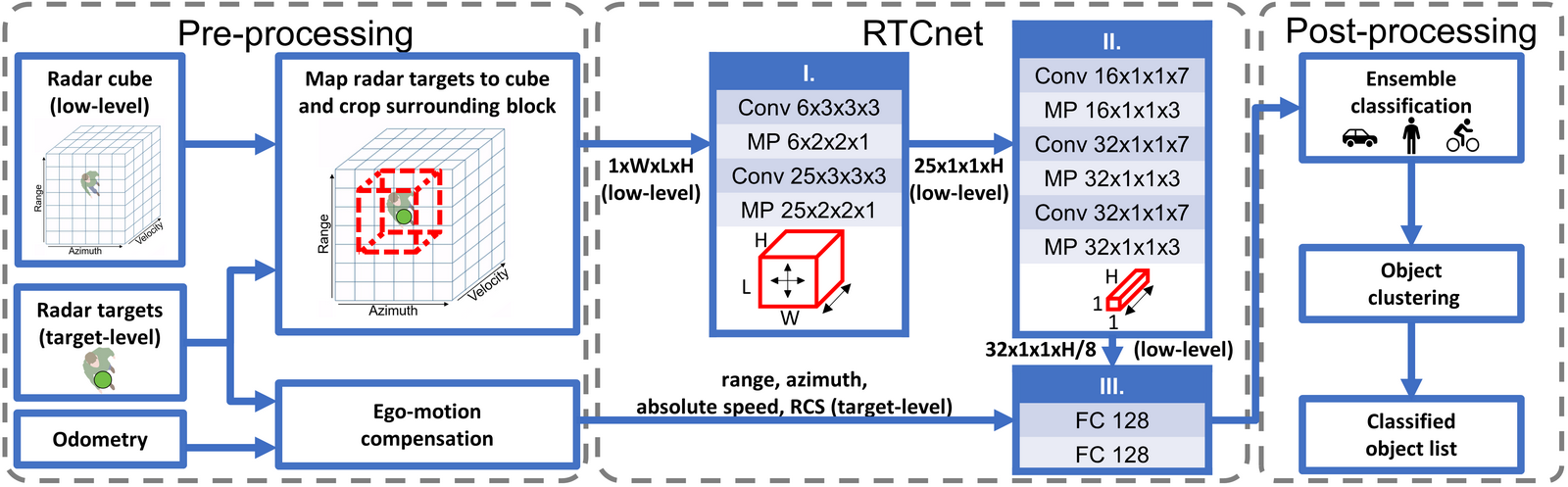}
	\caption{Our pipeline. 
		A block	around each radar target is cropped from radar cube.
		\LLTensemble has three parts.
		I. encodes range and azimuth dimensions. 
		II. extracts class information from the speed distribution. 
		III. provides scores based on II. and target-level features.
		Ensembling assigns a class label to each radar target. The class-specific clustering provides object proposals.
	}
	\label{fig:our_pipeline}
\end{figure*}
\section{PROPOSED METHOD}
\label{sec:method}
In this research, we combine the advantages of target-level (accurate range and azimuth estimation) and low-level data (more information in speed domain) by mapping the radar targets into the radar cube  and cropping a smaller block around it in all three dimensions (subsection \ref{subsec:preprocessing}). 
\LLTensemble classifies each target individually based on the fused low-level and target-level data. 
The network consists of three parts (subsection \ref{subsec:network}). 
The first encodes the data in spatial domains (range, azimuth) and grasps the surroundings' Doppler distribution. 
The second is applied on this output to extract class information from the distribution of speed.
Finally, the third part provides classifications scores by two fully connected layers (\rev{FC}). The output is either multi-class (\rev{one} score for each class) or binary. In the latter case, an ensemble voting (subsection \ref{subsec:ensembling}) step combines the result of several binary classifiers similarly to \cite{Scheiner2019}.
A class-specific clustering step (i.e. the radar targets' predicted class information is used) generates an object list output (subsection \ref{subsec:post-clustering}). See Fig. \ref{fig:our_pipeline} for an overview of our method. 
\rev{The software of our pipeline is available on our website\footnote{\url{https://github.com/tudelft-iv/RTCnet}}.}%
\subsection{Pre-processing} 
\label{subsec:preprocessing}
First, a single frame of radar targets and a single frame of the radar cube (low-level data) is fetched. 
Each radar target's speed is compensated for ego-motion similarly to \cite{Schumann2017}.
As we only address moving road users, radar targets with low compensated (absolute) velocity are considered as static and are filtered out. 
Then, corresponding target-level and low-level radar data are connected. 
That is, we look up each remaining dynamic radar target's corresponding range/azimuth/Doppler bins, i.e. a grid cell in the radar cube based on their reported range, azimuth and (relative) velocity ($ \rangehigh, \azimuthhigh, \velr $).
Afterwards, a 3D block of the radar cube is cropped around each radar target's grid cell with radius in range/azimuth/Doppler dimensions ($\rangecroprange, \azimuthcroprange, \dopplercroprange $). 
See "Pre-Processing" part on Fig. \ref{fig:our_pipeline}.
%
%

\subsection{Network}
\label{subsec:network}
\LLTensemble consists of three modules as seen on Fig. \ref{fig:our_pipeline}.
\subsubsection{Down-sample range and azimuth dimensions}
The first part's aim is to encode the radar target's spatial neighborhood's Doppler distribution into a tensor without extension in range or azimuth. 
In other words, it transforms the $1 \times \azimuthcroprange \times \rangecroprange \times \dopplercroprange $ sized data to a  \rev{$\ChannelNum \times 1 \times 1 \times \dopplercroprange $} sized tensor (sizes are given as $Channel \times Azimuth \times Range \times Doppler $), where \rev{$ \ChannelNum{} $ was chosen as $25$}.
To do this, it contains two 3D convolutions (Conv) with the kernel sizes of $6 \times 3\times3\times3$ and $25 \times 3\times3\times3$ (padding is $ 1 $). 
Both convolutional layers are followed by a maxpool (MP) layer with the kernel sizes of $ 6 \times 2 \times 2 \times 1 $ and $ 25 \times 2 \times 2 \times 1 $ with $ 0 $ padding to down-sample in the spatial dimensions. \looseness=-1
%

\subsubsection{Process Doppler dimension}
The second part of the network operates on the output of the first which is a $25 \times 1 \times 1 \times \dopplercroprange $ sized tensor.  
The aim of this module is to extract class information from the speed distribution around the target. 
To do this, we use three 1D convolutions along the Doppler dimension with the kernel size of $7$ and output channel sizes of $16, 32, 32$. Each convolution is followed by a maxpool layer with the kernel size of $3$ and stride of $2$, which halves the length of the input.
The output of the this module is a $32 \times 1 \times 1 \times \dopplercroprange/8$ block. 

\subsubsection{Score calculation}
The output of the second module is flattened and concatenated to the target-level features ($ \rangehigh, \azimuthhigh, \velr, \RCS $), and fed into the third one.
We use two fully connected layers with 128 nodes each to provide scores. The output layer has either four nodes (one for each class) for multi-class classification or two for binary tasks. In the latter case, ensemble voting is applied, see next subsection.

\subsection{Ensemble classifying}
\label{subsec:ensembling}
With four output nodes, it is possible to train the third module to perform multi-class classification directly.
We also implemented an ensemble voting system of binary classifiers (networks with two output nodes). That is, aside training a single, multi-class network, we followed
\cite{Scheiner2019} and trained One-vs-All (OvA) and One-vs-One (OvO) binary classifiers for each class (e.g. car-vs-all) and pair of classes (e.g. car-vs-cyclist), 10 in total. 
The final prediction scores depend on the voting of all the binary models. 
OvO scores are weighted by the summation of the corresponding OvA scores to achieve a more balanced result. 
\rev{Although we experimented with ensembling multi-class classifiers trained on bootstrapped training data as well, it yielded worse results.}
%

\subsection[Object Clustering]{\DMG{Object Clustering}}
\label{subsec:post-clustering}

The output of the network (or voting) is a predicted \DMG{class} label for each target \DMG{individually}. \DMG{To obtain} proposals \DMG{for object detection}, we cluster the classified radar targets with DBSCAN incorporating the predicted class information, i.e. radar targets with \textit{bike/pedestrian/car} predicted labels are clustered in separate steps. 
As metric, we used a spatial threshold  \rev{$\spatialRadius$} on the Euclidean distance in the $ x, y $ space (2D Cartesian spatial position), and a separate speed threshold \rev{$\velocityRadius$} in velocity dimension (\baselineProphet, \cite{Scheiner2019a}, \cite{Schubert2015Clustering}). 
The advantage of clustering each class separately is that no universal parameter set is needed for DBSCAN. Instead, we can use different parameters for each class, e.g. larger radius for cars and small ones for pedestrians (Fig. \ref{fig:Clustering_errors}, A and B).
Furthermore, swapping the clustering and classification step makes it possible to consider objects with a single \DMG{reflection}, e.g. setting $\MinPoints$ to one for pedestrian labeled radar targets (Fig. \ref{fig:Clustering_errors}, C).
A possible drawback is that if a subset of an object's reflections are misclassified (e.g. a car with multiple targets, most labeled \textit{car} and some as \textit{cyclist}), the falsely classified targets (i.e. the \textit{cyclist} ones) will be mistakenly clustered into a separate object. 
To address this, we perform a filtering on the produced object proposals, calculating their \rev{spatial}, (radial) velocity, and class score distribution distances (scores are handled as 4D vector, and we take their Euclidean distance after normalization). \rev{If two clusters have different classes and are close enough in all dimensions} \DMG{(cf. parameters in Sect.} \ref{subsec:implementation}), \rev{ we merge the smaller class to the larger (i.e. pedestrians to cyclists and cars, cyclists to cars) given that the cluster from the larger class has more radar targets.}
\section{DATASET}

Our real-world dataset contains $ \sim 1 $ hour of  driving in urban environment with our demonstrator vehicle \cite{Prius2019}. 
We recorded both the target-level and low-level output of our radar, a Continental 400 series mounted behind the front bumper. 
We also recorded the output of a stereo camera {($1936 \times 1216$ px)} mounted on the wind-shield, 
and the ego-vehicle's odometry (filtered location and ego-speed).

Annotation was fetched automatically from the camera sensor using the Single Shot Multibox Detector (SSD) \cite{Liu2016} trained on the EuroCity Persons dataset \cite{Braun2019}. Distance is estimated by projecting each bounding box into the stereo point-cloud computed by the Semi-Global Matching algorithm (SGM) \cite{Hirschmuller2008}, and taking the median distance of the points inside each.
In a second iteration, we manually corrected mislabeled ground truth, e.g. cyclist annotated as pedestrian.
The training set contains more than $30/15/9 \times 10^3$ pedestrian/cyclist/car instances respectively (one object may appear on several frames), see 
Table \ref{table:dateset}. 
Fig. \ref{fig:range_dependent_results} shows the distribution of radar targets in the training set distance-wise.
\begin{table} []
	\centering
	\begin{tabular}{@{}llll@{}}
		\toprule
		                                          & Pedestrians & Bikers &  Cars  \\ \midrule
		Number of instances                       &    31300    & 15290  &  9362  \\
		Number of radar targets                   &    63814    & 45804  & 30906  \\
		Avg. number of radar targets per instance &    2.04     &  3.00  &  3.30  \\
		Instances with only one radar target      &    12990    &  3526  &  2878  \\ \midrule
		{Ratio of instances with one radar target}                    &  {41.5\%}   & 18.8\% & 37.6\% \\ \bottomrule
	\end{tabular}
	\caption{Number of instances from each class in our training set. Many road users have only one radar reflection, which is not enough to extract meaningful statistical features. } 
	\label{table:dateset}
\end{table}
To further extend our training dataset, we augmented the data by mirroring the radar frames and adding a zero-mean, $ 0.05 $ std Gaussian noise to the normalized $ \rangehigh$ and $\velr$ features.	
Training and testing sets are from two independent driving (33 and 31 minutes long) which took place on different days and routes. 
\rev{Validation set is a 10\% split of training dataset after shuffling.}

\section{EXPERIMENTS}


We compared our proposed method, \LLTensemble \rev{ with binary bagging (from now on, referred to as} \LLTensemble) to two baselines in two experiments to examine their radar target classification and object detection capabilities.  

In the first experiment, we examined their performance in classification task, using a target-wise metric, i.e. a true positive is a correctly classified target \cite{Schumann2018}. 
For cluster-wise methods (the baselines) the predicted label of a cluster is assigned to each radar target inside following \cite{Schumann2018}.
Furthermore, we also performed an ablation study to see how different features benefit our method in this classification (\rev{adaptation} in brackets).
\LLTmulticlass is a single, multi-class network to see if ensembling is beneficial.
\LLTnoRCS is identical to \LLTensemble, but the RCS target-level feature is removed to examine its importance. 
Similarly, in \LLTnospeed the absolute speed of the targets is unknown to the networks, only the relative speed distribution (in the low-level data) is given.
Finally, \LLThigh is a significantly modified version as it only uses target-level features. That is, the first and second convolutional parts are skipped, and the radar targets are fed to the third fully connected part directly. Note that in contrast to \LLTnospeed, \LLThigh has access to the absolute speed of the target, but lacks the relative speed distribution. \DMG{Object} clustering is skipped in the first experiment. \looseness=-1 

In the second experiment, we compare the methods in object detection task, examining our whole pipeline, including the \rev{object} clustering step. 
Predictions and annotations are compared by their intersection and union calculated in number of targets, see Fig. \ref{fig:iou}.
A true positive is a prediction which has an Intersection Over Union (IoU) bigger than or equal to $0.5$ with an annotated object. Further detections of the same ground truth object count as false positives.

All presented results were measured on moving radar targets to focus on moving road users.
\begin{figure}
	\centering
	\includegraphics[width=\linewidth]{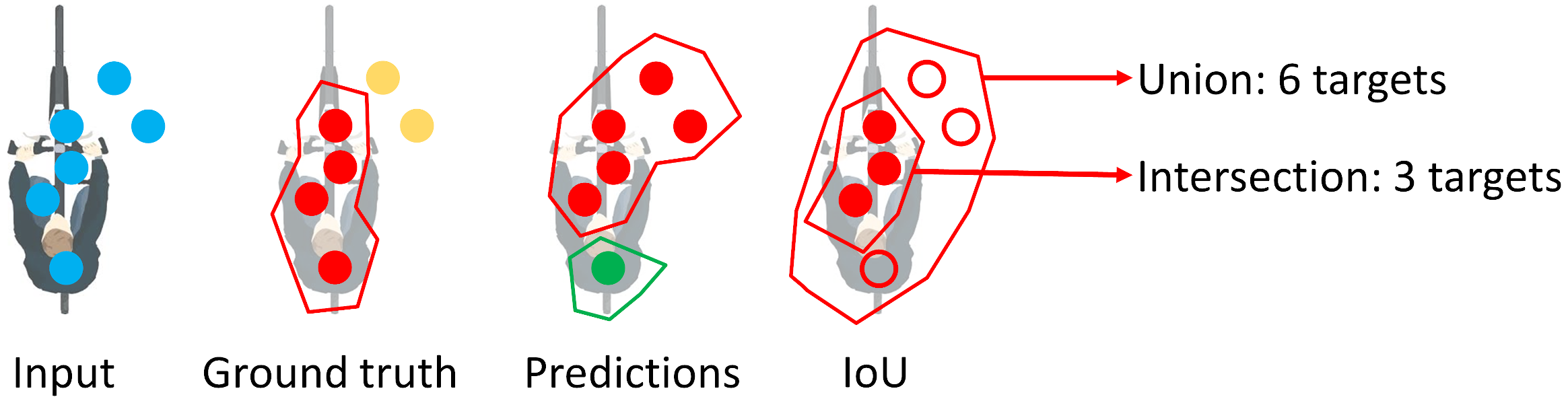}
	\caption{Object-level metric. Intersection and Union are defined by number of radar targets. ${ \frac{Intesection}{Union} \geq 0.5}$ counts as a true positive. In this example, there is a true positive cyclist and a false positive pedestrian detection.}
	\label{fig:iou}
\end{figure}

\subsection{Baselines}
\label{subsec:baselines}
We selected \baselineSchumann as baseline because it is the only multi-object, multi-class detection method found with \DMG{small} latency, see Table \ref{tab:overview_of_methods}. As no other research handled multiple classes, we selected \baselineProphet as our second baseline, which is a single-class pedestrian detector, but the negative training and testing set contained cars, dogs, and cyclists. 
We re-implemented their full pipeline (DBSCAN clustering and cluster-wise classification) and trained their algorithms with our training set.
Optimal DBSCAN parameters are sensor specific (depending on density, resolution, etc.), thus we optimized 
the threshold in spatial dimensions \rev{$\spatialRadius$} ($ 0.5~m-1.5~m$, step size $0.1~m$) and 
the threshold in velocity  \rev{$\velocityRadius$}  ($ 0.5-1.5~m/s $, step size $ 0.1~m/s$)
on our validation set for both baselines independently. We used the same metric as in our object clustering. 
%
Both baselines have features describing the number of static radar targets in the cluster. We also searched for an optimal speed threshold $\MinSpeed$ ($ 0-0.5~m/s $, step size $ 0.1~m/s$) for both to define these static radar targets. 
All reported results for baselines were reached by using their optimal settings, see Table \ref{tab:parameters}. $ \MinPoints $ was set to two as in \baselineProphet (increasing it further would exclude almost all pedestrians, see Table \ref{table:dateset}). In \baselineSchumann the authors used manually corrected clusters (i.e. separating objects falsely merged by DBSCAN) to focus on the classification. We did not correct them to examine real-life application possibilities. 
We implemented a Random Forest classifier with 50 trees for both baselines, as \baselineProphet reported it to be the best for their features. \baselineSchumann also tested LSTM, but used several frames aggregated as input.

\begin{table}
	\centering
	\begin{tabular}{@{}lcccc@{}}  \toprule
	Method                & \rev{$\spatialRadius$} & \rev{$\velocityRadius$} & $\MinPoints$ & $\MinSpeed$ \\ \midrule
	\baselineProphet      &       \rev{$1.2~m$}        &      \rev{$1.3~m/s$}      &      2       &  $0.4~m/s$  \\
	\baselineSchumann     &       \rev{$1.3~m$}        &      \rev{$1.4~m/s$}      &      2       &  $0.4~m/s$  \\
	Class-specific (peds.)    &       $0.5~m$        &      $2.0~m/s$      &      1       &  $-$  \\
	Class-specific (cyclists) &       $1.6~m$        &      $1.5~m/s$      &      2       &  $-$  \\
	Class-specific (cars)     &       $4.0~m$        &      $1.0~m/s$      &      3       &  $-$  \\ \bottomrule
	\end{tabular}%
	\caption{Optimized DBSCAN parameters for the two baselines, and for our class-specific clustering for each class.}
	\label{tab:parameters}%
\end{table}%

\subsection{Implementation}
\label{subsec:implementation}

We set $\rangecroprange = \azimuthcroprange = 5$, $\dopplercroprange=32$ as the size of the cropped block. 
Speed threshold to filter out static objects \rev{is a sensor specific parameter} and was set to $ 0.3~m/s$ \rev{based on empirical evidence}.
Table \ref{tab:parameters} shows the DBSCAN parameters for both baselines and for our class-specific clustering step. \rev{The thresholds to merge clusters during object clustering were set to {$1~m$} spatially, {$0.6$} for scores, {$2~m/s$} for pedestrian to cyclist, and {$1.2~m/s$} for pedestrian/cyclist to car merges.}

We normalized the data to be zero-mean and have a standard deviation of 1 feature-wise for $\rangehigh, \azimuthhigh, \velr, \RCS$, and for the whole radar cube. 
At inference values calculated from training data are used.
We used PyTorch \cite{paszke2017automatic} for training with a cross-entropy loss (after softmax) in 10 training epochs.
Inference time is $\sim0.04~s$ on a {high-end PC} (Nvidia TITAN V GPU, Intel Xeon E5-1650 CPU, 64 GB RAM), including all moving radar targets, the $ 10 $ binary classifiers and the ensembling. 



\begin{figure*}
	\centering
	\hfill
	\begin{subfigure}{0.125\linewidth}
		\centering
		\adjincludegraphics[width=0.9\linewidth, height = 1.61\linewidth]{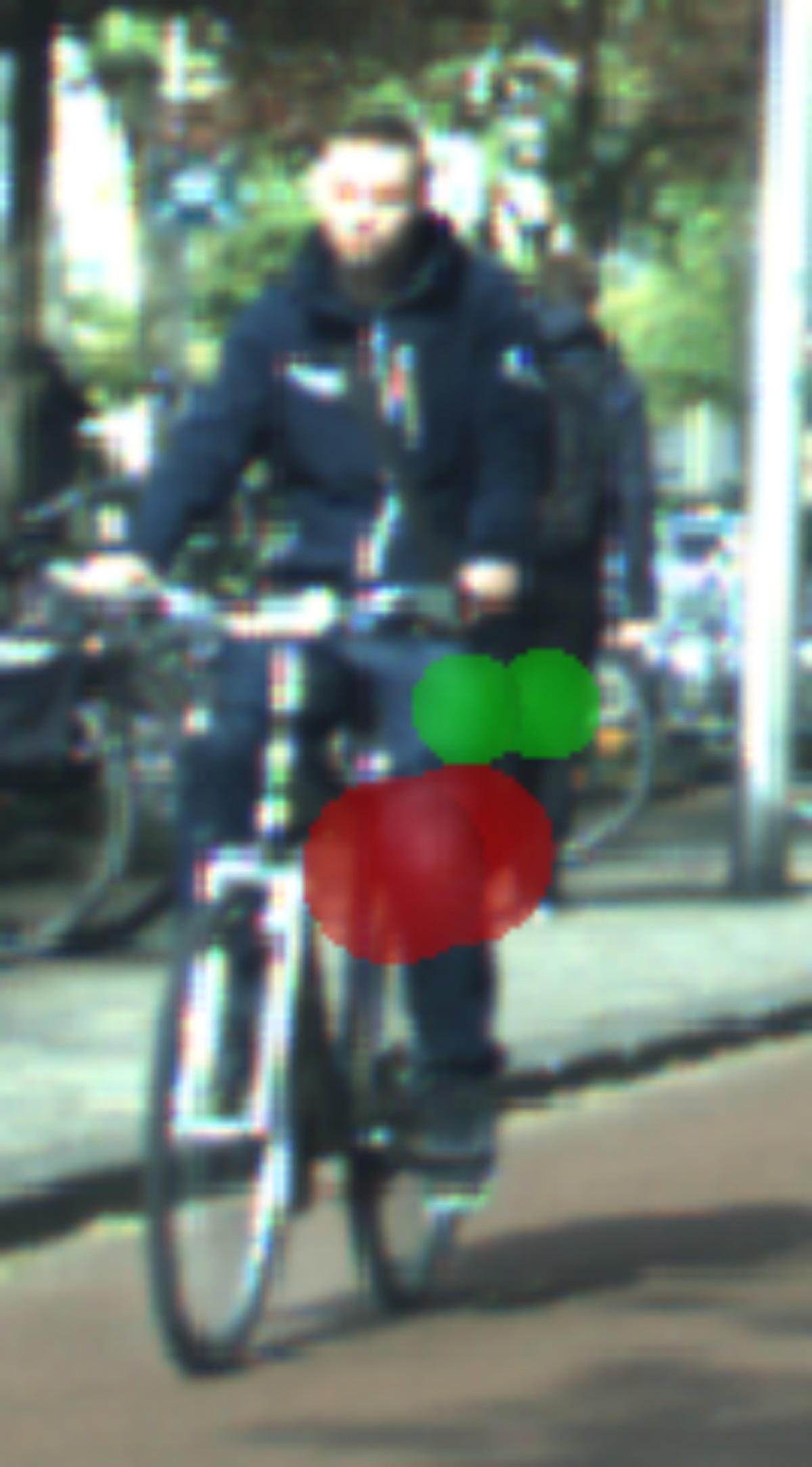}
		\caption{}
		\label{subfig:sample_1}
	\end{subfigure}%
	\hfill
	\begin{subfigure}{0.125\linewidth}
		\centering
		\adjincludegraphics[width=0.9\linewidth, height = 1.61\linewidth]{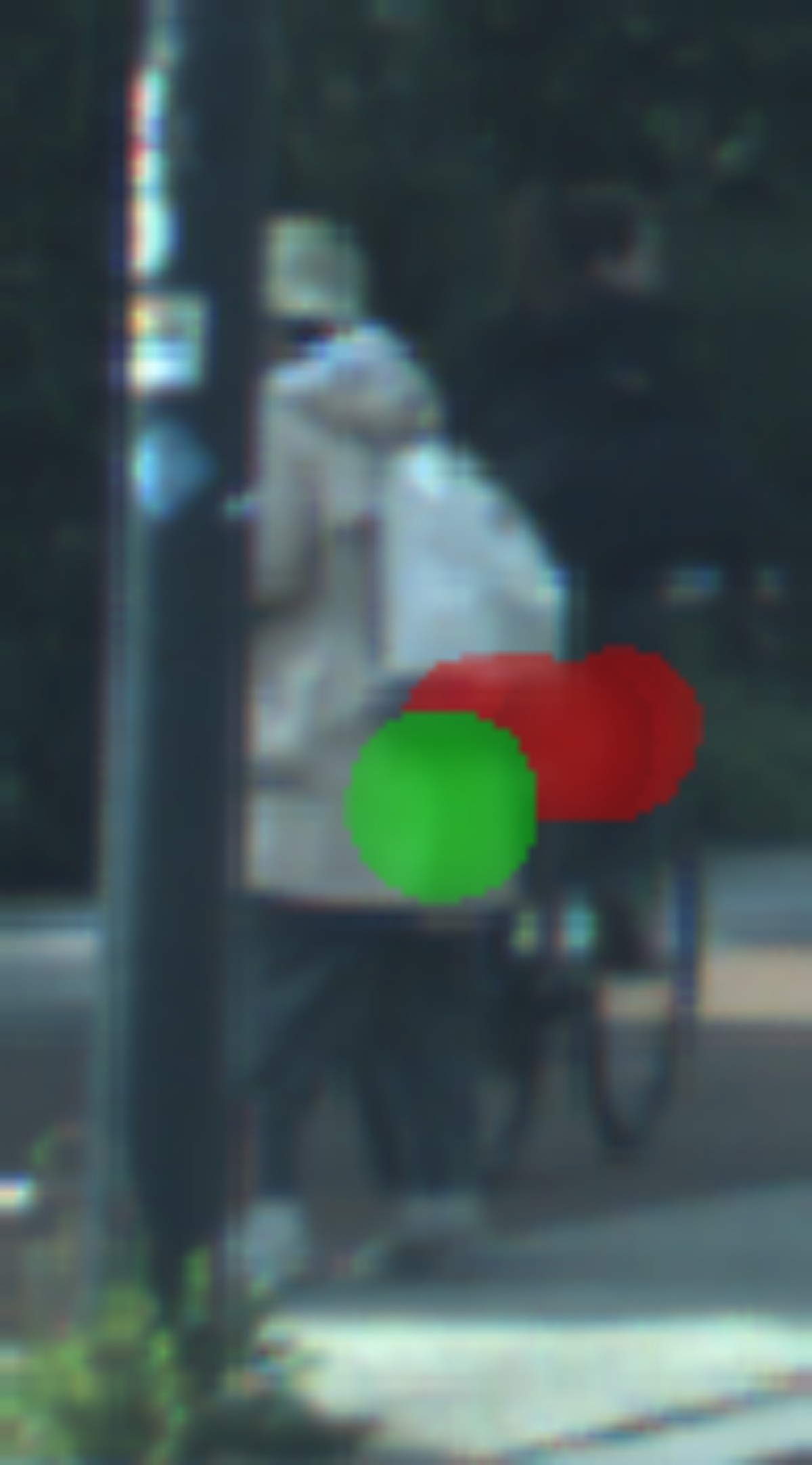}
		\caption{}
		\label{subfig:sample_2}
	\end{subfigure}%
	\hfill
	\begin{subfigure}{0.125\linewidth}
		\centering
		\adjincludegraphics[width=0.9\linewidth, height = 1.61\linewidth]{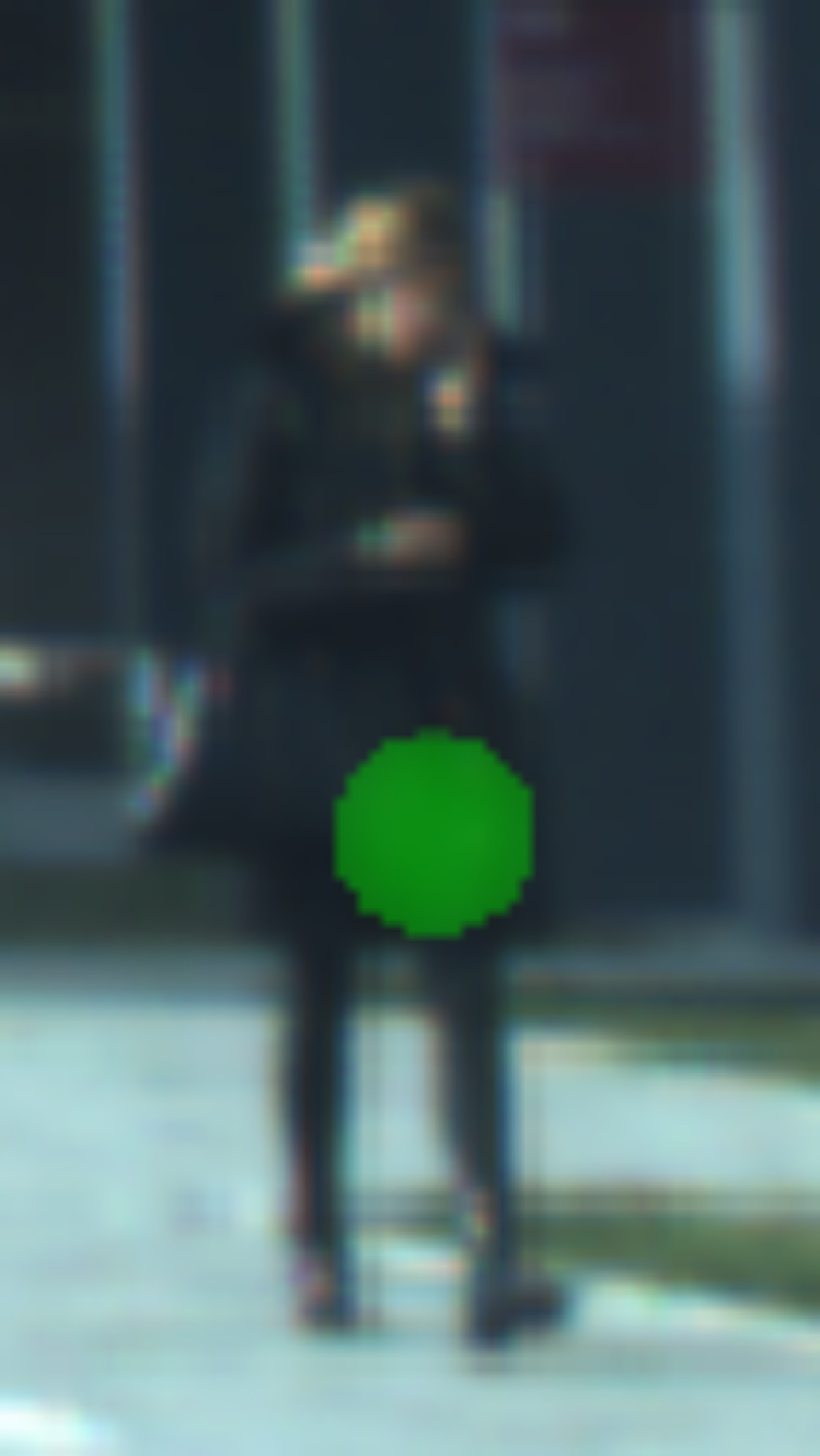}
		\caption{}
		\label{subfig:sample_3}
	\end{subfigure}%
	\hfill
	\begin{subfigure}{0.125\linewidth}
		\centering
		\adjincludegraphics[width=0.9\linewidth, height = 1.61\linewidth]{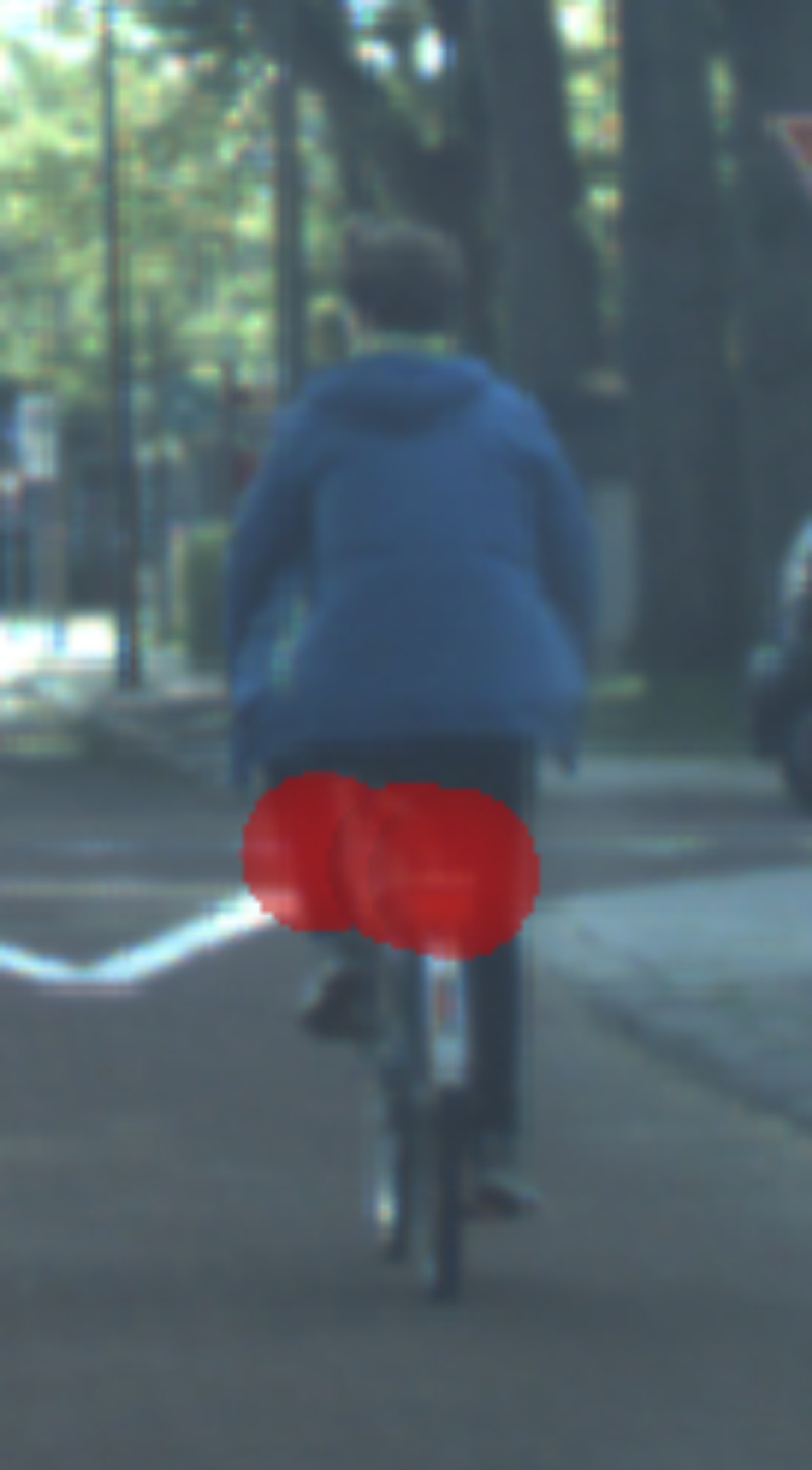}
		\caption{}
		\label{subfig:sample_4}
	\end{subfigure}%
	\hfill
	\begin{subfigure}{0.125\linewidth}
		\centering
		\adjincludegraphics[width=0.9\linewidth, height = 1.61\linewidth]{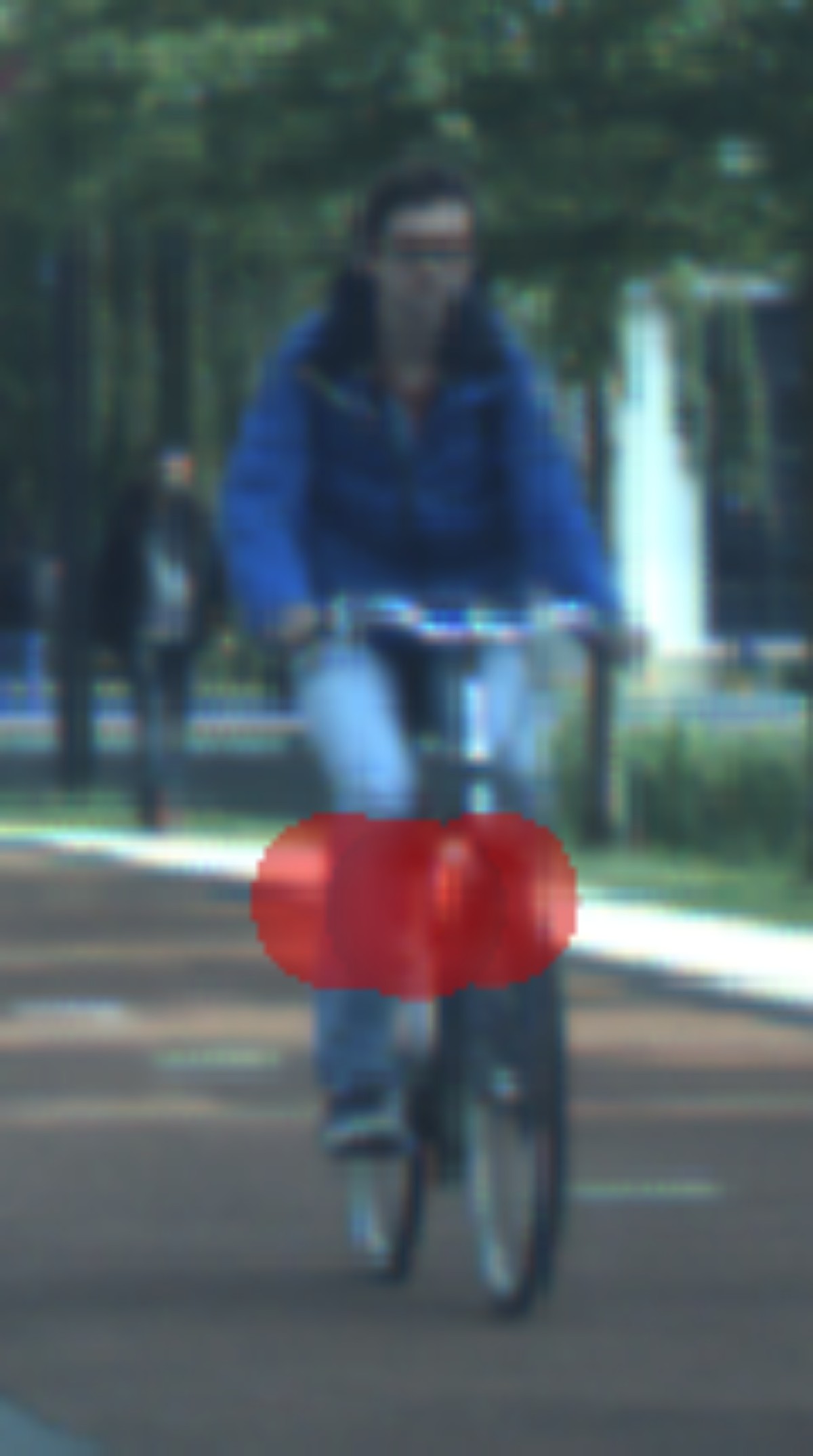}
		\caption{}
		\label{subfig:sample_5}
	\end{subfigure}%
	\hfill
	\begin{subfigure}{0.125\linewidth}
		\centering
		\adjincludegraphics[width=0.9\linewidth, height = 1.61\linewidth]{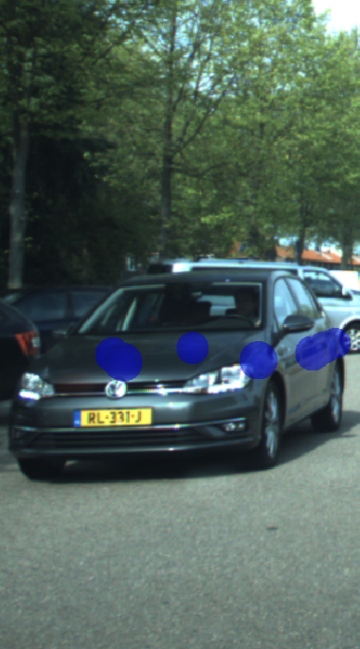}
		\caption{}
		\label{subfig:sample_6}
	\end{subfigure}%
	\hfill
	\begin{subfigure}{0.125\linewidth}
		\centering
		\adjincludegraphics[width=0.9\linewidth, height = 1.61\linewidth]{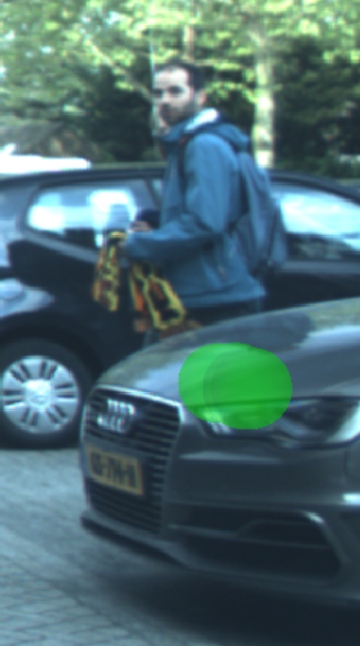}
		\caption{}
		\label{subfig:sample_7}
	\end{subfigure}%
	\hfill
	\begin{subfigure}{0.125\linewidth}
		\centering
		\adjincludegraphics[width=0.9\linewidth, height = 1.61\linewidth]{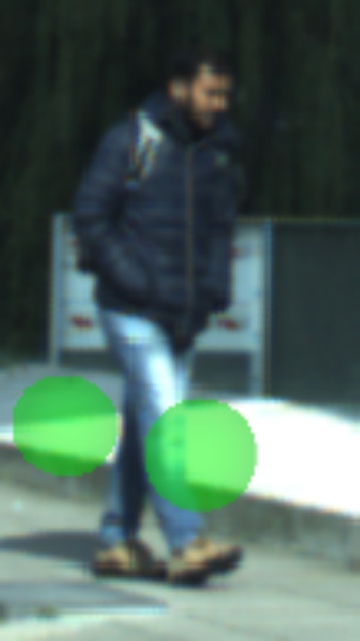}
		\caption{}
		\label{subfig:sample_8}
	\end{subfigure}%
	\caption{Examples of correctly classified radar targets by \LLTensemble, projected to image plane. Radar targets  with pedestrian/cyclist/car labels are marked by green/red/blue. Static objects and the class \textit{other} are not shown.
	} 
	\label{fig:sample_detections}
\end{figure*}

\begin{figure}
	\begin{subfigure}{0.25\linewidth}
		\centering
		\adjincludegraphics[width=0.9\linewidth, Clip = {0\width} {0.1\height}  {0\width} {0.1\height}]{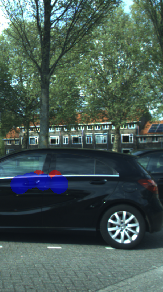}
		\caption{}
		\label{subfig:error_2}
	\end{subfigure}%
	\begin{subfigure}{0.25\linewidth}
		\centering
		\adjincludegraphics[width=0.9\linewidth, Clip = {0\width} {0.1\height}  {0\width} {0.1\height}]{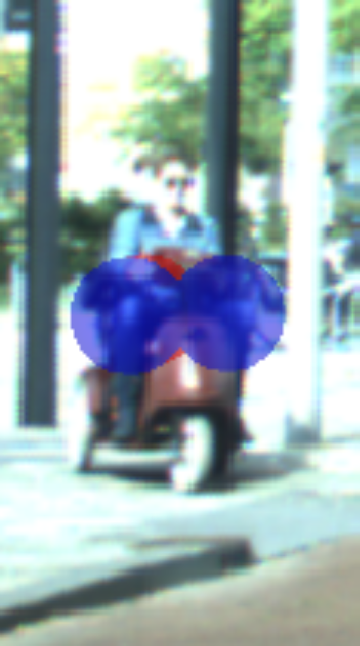}
		\caption{}
		\label{subfig:error_5}
	\end{subfigure}%
	\begin{subfigure}{0.25\linewidth}
		\centering
		\adjincludegraphics[width=0.9\linewidth, Clip = {0\width} {0.1\height}  {0\width} {0.1\height}]{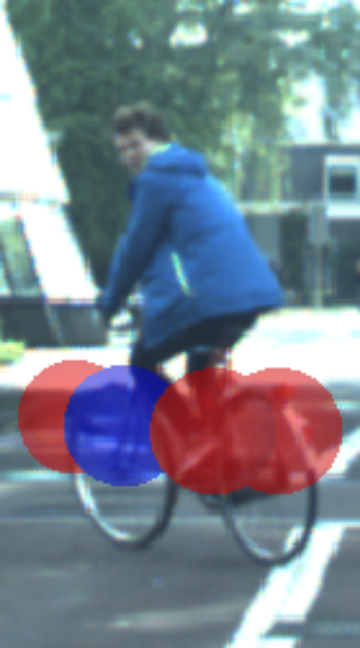}
		\caption{}
		\label{subfig:error_6}
	\end{subfigure}%
	\begin{subfigure}{0.25\linewidth}
		\centering
		\adjincludegraphics[width=0.9\linewidth, Clip = {0\width} {0.1\height}  {0\width} {0.1\height}]{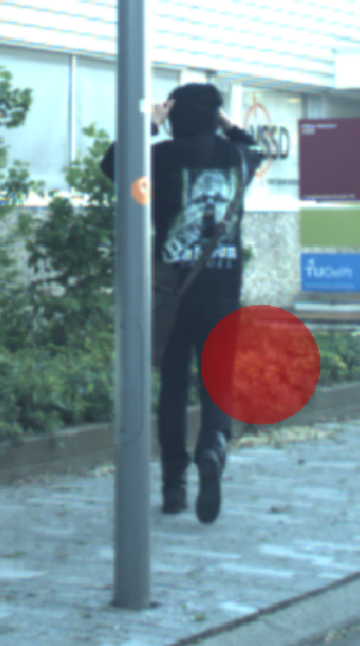}
		\caption{}
		\label{subfig:error_7}
	\end{subfigure}%
	\caption{Examples of radar targets misclassified by \LLTensemble, caused by: flat surfaces acting as mirrors and creating ghost targets \subref{subfig:error_2}, unusual vehicles \subref{subfig:error_5}, partial misclassification of an objects' reflections \subref{subfig:error_6}, and strong reflections nearby \subref{subfig:error_7}.} 
	\label{fig:sample_errors}
\end{figure}

\subsection{Results}
\begin{table}
	\centering
	\begin{tabular}{@{}lccccc@{}}
		\toprule
		Method            &     Pedestrian      &       Cyclist       &      Car       &        Other        &        Avg.         \\ \midrule
		\baselineProphet  &        0.61         &     \rev{0.58}      &   \rev{0.34}   &     \rev{0.91}      &     \rev{0.61}      \\
		\baselineSchumann &     \rev{0.67}      & \rev{\textbf{0.68}} &   \rev{0.46}   & \rev{\textbf{0.92}} &     \rev{0.68}      \\
		\LLThigh          &     \rev{0.56}      &     \rev{0.63}      &   \rev{0.33}   &        0.90         &        0.61         \\
		\LLTnospeed       &     \rev{0.66}      &     \rev{0.63}      &     {0.36}     &        0.91         &     \rev{0.64}      \\
		\LLTnoRCS         &   \textbf{ 0.71}    &     \rev{0.66}      &   \rev{0.48}   &        0.91         &        0.69         \\
		\LLTmulticlass    &     \rev{0.67}      &     \rev{0.65}      &   \rev{0.47}   &        0.89         &     \rev{0.67}      \\
		\LLTensemble      & \rev{\textbf{0.71}} &     \rev{0.67}      & \textbf{ 0.50} &    \textbf{0.92}    & \rev{\textbf{0.70}} \\ \bottomrule
	\end{tabular}%
	\caption{Target-wise F1 scores per class (best in bold). \LLTensemble outperforms the baselines on average. The ablation study shows benefits of ensembling and using low-level data.}
	\label{tab:F1results}%
\end{table}%
\subsubsection{Target classification}
We present the results of the target classification experiment in Table \ref{tab:F1results}. Target-wise F1 scores for all classes and their macro-average are given for each method. 
\LLTensemble outperformed the two cluster-wise baselines reaching an average F1 score of \rev{0.70}. \baselineSchumann has slightly better results on \rev{\textit{cyclists}} than \LLTensemble (0.68 vs 0.67), but performed significantly worse on \textit{pedestrians} (0.67 vs 0.71) and \textit{cars} (0.46. vs 0.50).
The ablation study showed that removing each feature 
yields worse results than the complete pipeline, but the one without reflectivity information (\LLTnoRCS) comes close with an average of 0.69. 
Removing the low-level features (\LLThigh) decreased the performance significantly to an average of 0.61.
The multi-class (single) network \LLTmulticlass outperforms the baselines on the \textit{car} class, but performs worse on \textit{cyclists}. Ensemble voting brings significant improvement on all classes.
Example of correct and incorrect target classifications are shown on Fig. \ref{fig:sample_detections} and \ref{fig:sample_errors} for all road user classes.
On Fig. \ref{fig:range_dependent_results} we show how the classification performance (target-wise F1 score) changes over distance (with $ 5~m $ bins) for each class, along with the number of radar targets in the training set.
Although most annotation fall into the $ 5-20~m$ range, the network performs reasonably beyond that distance, especially for the larger objects (\textit{cyclist, car}).
We trained One-vs-All classifiers both for \LLTensemble and \baselineSchumann for each road user class, and plotted their performance on receiver operating characteristic (ROC) curves on Fig. \ref{fig:roc_curves}.
The varied threshold is cluster-wise for \baselineSchumann and target-wise for \LLTensemble. Our method has a larger area under the curve of all classes.

\begin{figure}
	\centering{
	\vspace{-14pt}
	\scalebox{1}{
\begin{tikzpicture}
\pgfplotsset{every x tick label/.append style={font=\small, yshift=0.5ex}}
\pgfplotsset{every y tick label/.append style={font=\small, xshift=0.5ex}}
\pgfplotsset{set layers}
\pgfplotsset{compat=1.11,
	/pgfplots/ybar legend/.style={
		/pgfplots/legend image code/.code={%
			\draw[##1,/tikz/.cd,yshift=-0.25em]
			(0cm,0cm) rectangle (3pt,0.8em);},
	},
}
\begin{axis}[
scaled ticks=false, tick label style={/pgf/number format/fixed},
scale only axis=true,
height=0.45\linewidth,
width = 0.77\linewidth,
legend cell align={left},
legend style={at={(0.03,0.97)}, anchor=north west, draw=white!80.0!black},
tick align=center,
tick pos=left,
x grid style={white!69.01960784313725!black},
xmin=-0.945, xmax=9.945,
xtick style={color=white},
x tick label style={rotate=90},
xticklabels={},
y grid style={white!69.01960784313725!black},
ylabel={Number of targets},
ymin=0, ymax=27000,
ytick style={color=black},
y tick label style={rotate=90}
]
\draw[fill=green!50.0!black,draw opacity=0] (axis cs:-0.45,0) rectangle (axis cs:-0.15,534);
\addlegendimage{ybar,ybar legend,fill=green!50.0!black,draw opacity=0};

\draw[fill=green!50.0!black,draw opacity=0] (axis cs:0.55,0) rectangle (axis cs:0.85,9974);
\draw[fill=green!50.0!black,draw opacity=0] (axis cs:1.55,0) rectangle (axis cs:1.85,23524);
\draw[fill=green!50.0!black,draw opacity=0] (axis cs:2.55,0) rectangle (axis cs:2.85,12536);
\draw[fill=green!50.0!black,draw opacity=0] (axis cs:3.55,0) rectangle (axis cs:3.85,5884);
\draw[fill=green!50.0!black,draw opacity=0] (axis cs:4.55,0) rectangle (axis cs:4.85,2746);
\draw[fill=green!50.0!black,draw opacity=0] (axis cs:5.55,0) rectangle (axis cs:5.85,1456);
\draw[fill=green!50.0!black,draw opacity=0] (axis cs:6.55,0) rectangle (axis cs:6.85,580);
\draw[fill=green!50.0!black,draw opacity=0] (axis cs:7.55,0) rectangle (axis cs:7.85,184);
\draw[fill=green!50.0!black,draw opacity=0] (axis cs:8.55,0) rectangle (axis cs:8.85,54);
\draw[fill=red,draw opacity=0] (axis cs:-0.15,0) rectangle (axis cs:0.15,1412);
\addlegendimage{ybar,ybar legend,fill=red,draw opacity=0};

\draw[fill=red,draw opacity=0] (axis cs:0.85,0) rectangle (axis cs:1.15,8446);
\draw[fill=red,draw opacity=0] (axis cs:1.85,0) rectangle (axis cs:2.15,8530);
\draw[fill=red,draw opacity=0] (axis cs:2.85,0) rectangle (axis cs:3.15,7660);
\draw[fill=red,draw opacity=0] (axis cs:3.85,0) rectangle (axis cs:4.15,5180);
\draw[fill=red,draw opacity=0] (axis cs:4.85,0) rectangle (axis cs:5.15,4452);
\draw[fill=red,draw opacity=0] (axis cs:5.85,0) rectangle (axis cs:6.15,2680);
\draw[fill=red,draw opacity=0] (axis cs:6.85,0) rectangle (axis cs:7.15,1672);
\draw[fill=red,draw opacity=0] (axis cs:7.85,0) rectangle (axis cs:8.15,668);
\draw[fill=red,draw opacity=0] (axis cs:8.85,0) rectangle (axis cs:9.15,298);
\draw[fill=blue,draw opacity=0] (axis cs:0.15,0) rectangle (axis cs:0.45,712);
\addlegendimage{ybar,ybar legend,fill=blue,draw opacity=0};

\draw[fill=blue,draw opacity=0] (axis cs:1.15,0) rectangle (axis cs:1.45,4748);
\draw[fill=blue,draw opacity=0] (axis cs:2.15,0) rectangle (axis cs:2.45,2642);
\draw[fill=blue,draw opacity=0] (axis cs:3.15,0) rectangle (axis cs:3.45,2638);
\draw[fill=blue,draw opacity=0] (axis cs:4.15,0) rectangle (axis cs:4.45,2048);
\draw[fill=blue,draw opacity=0] (axis cs:5.15,0) rectangle (axis cs:5.45,1346);
\draw[fill=blue,draw opacity=0] (axis cs:6.15,0) rectangle (axis cs:6.45,1500);
\draw[fill=blue,draw opacity=0] (axis cs:7.15,0) rectangle (axis cs:7.45,1548);
\draw[fill=blue,draw opacity=0] (axis cs:8.15,0) rectangle (axis cs:8.45,1310);
\draw[fill=blue,draw opacity=0] (axis cs:9.15,0) rectangle (axis cs:9.45,1224);
\end{axis}

\begin{axis}[
axis y line*=right,
tick align=center,
scale only axis=true,
height=0.45\linewidth,
width = 0.77\linewidth,
legend cell align={left},
legend style={draw=white!80.0!black,
			  fill = none,
			  font=\footnotesize},
legend image post style={scale=0.4},
xmin=-0.945, xmax=9.945,
xlabel={Range (m)},
xtick style={color=black},
xtick pos=left,
x tick label style={rotate=45},
xtick={0,1,2,3,4,5,6,7,8,9,10},
xticklabels={0-5,5-10,10-15,15-20,20-25,25-30,30-35,35-40,40-45,45-50,50-55},
y grid style={white!69.01960784313725!black},
ylabel={F1 score},
ymin=0, ymax=0.85,
ytick style={color=black},
y tick label style={rotate=90}
]
\addplot [only marks, draw=green!50.0!black, fill=green!50.0!black, colormap/viridis, forget plot, mark size=1pt]
table{%
x                      y
0 0.382978723404255
1 0.76671379519289
2 0.793141147361758
3 0.714038461538462
4 0.680773028635915
5 0.611923509561305
6 0.442079410560786
7 0.230051813471503
8 0
9 0
};
\addplot [only marks, draw=red, fill=red, colormap/viridis, forget plot, mark size=1pt]
table{%
x                      y
0 0.504253147329023
1 0.599251973957612
2 0.735425627049355
3 0.738880484114977
4 0.722864538395168
5 0.71351845503371
6 0.694406754108247
7 0.628814213982233
8 0.258724988366682
9 0.00802139037433155
};
\addplot [only marks, draw=blue, fill=blue, colormap/viridis, forget plot, mark size=1pt]
table{%
x                      y
0 0.193877551020408
1 0.468431771894094
2 0.542475591964987
3 0.551612428213812
4 0.556137289121582
5 0.506473329880891
6 0.520470829068577
7 0.505992010652463
8 0.51694646816598
9 0.557259713701432
};
\addplot [semithick, green!50.0!black]
table {%
0 0.382978723404255
1 0.76671379519289
2 0.793141147361758
3 0.714038461538462
4 0.680773028635915
5 0.611923509561305
6 0.442079410560786
7 0.230051813471503
8 0
9 0
};
\addlegendentry{Ped.}
\addplot [semithick, red]
table {%
0 0.504253147329023
1 0.599251973957612
2 0.735425627049355
3 0.738880484114977
4 0.722864538395168
5 0.71351845503371
6 0.694406754108247
7 0.628814213982233
8 0.258724988366682
9 0.00802139037433155
};
\addlegendentry{Biker}
\addplot [semithick, blue]
table {%
0 0.193877551020408
1 0.468431771894094
2 0.542475591964987
3 0.551612428213812
4 0.556137289121582
5 0.506473329880891
6 0.520470829068577
7 0.505992010652463
8 0.51694646816598
9 0.557259713701432
};
\addlegendentry{Car}
\end{axis}

\end{tikzpicture}}}
	\caption{Target-wise F1 scores (lines) and number of targets in training set (bars) in function of distance from ego-vehicle.}
	\label{fig:range_dependent_results}
\end{figure}
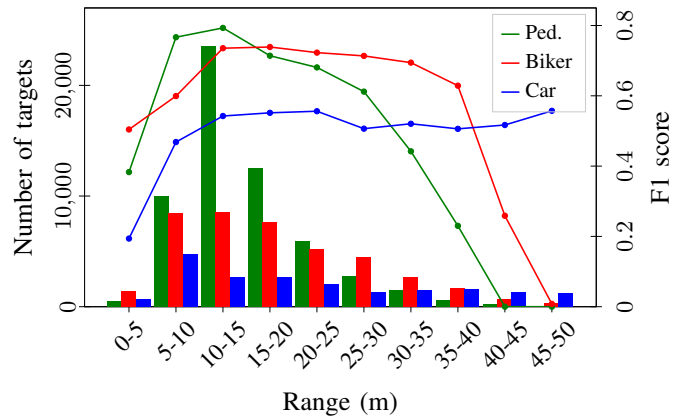
\begin{figure}
	\vspace{-10pt}
	\scalebox{1}{\input{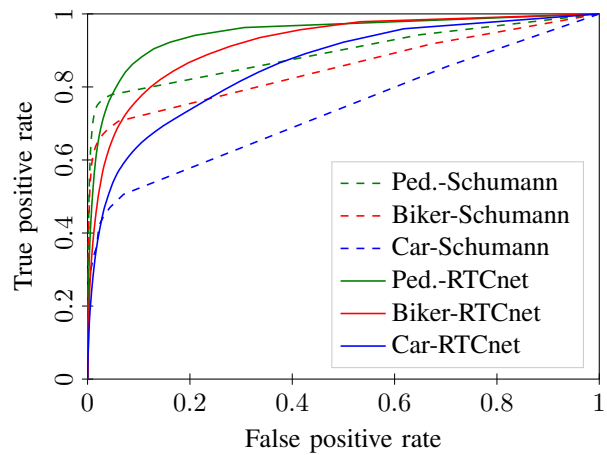}}
	\caption{ROC curves of road user classes by our method and \baselineSchumann. Each curve is calculated by changing the decision threshold of a One-vs-All binary classifier. 
	}
	\label{fig:roc_curves}
\end{figure}
\subsubsection{Object detection}
The results of our second experiment are shown in Table \ref{tab:F1objresults}. 
\LLTensemble reached \rev{slightly worse results on \textit{cyclists} than} \baselineSchumann \xspace{\rev{(0.59 vs 0.60)}, but significantly outperformed it on \textit{pedestrians} (0.61 vs 0.54), \textit{cars} (0.47 vs 0.31), and in average (0.56 vs 0.48).}
Fig. \ref{fig:baseline_vs_us} shows how \baselineSchumann and \LLTensemble handled two real-life cases from Fig. \ref{fig:Clustering_errors}.
Examples for both correct and incorrect object detections by \LLTensemble are shown on Fig. \ref{fig:cl_sample_errors}.
A link to a video of our results can be found on our website\footnote{
\url{{http://intelligent-vehicles.org/publications/}}}.
\begin{table}
	\centering
	\begin{tabular}{@{}lcccc@{}}
		\toprule
		              &     Pedestrian      &       Cyclist       &         Car         &        Avg.         \\ \midrule
		\baselineProphet    &        0.48         &     \rev{0.50}      &     \rev{0.23}      &     \rev{0.40}      \\
		\baselineSchumann   &     \rev{0.54}      & \rev{\textbf{0.60}} &     \rev{0.31}      &     \rev{0.48}      \\
		\LLTensemble (ours) & \rev{\textbf{0.61}} &    \rev{{0.59}}     & \rev{\textbf{0.47}} & \rev{\textbf{0.56}} \\ \bottomrule
	\end{tabular}%
	\caption{{F1 scores object-wise} (best score in bold). {\LLTensemble outperforms the baselines on average.}}
	\label{tab:F1objresults}%
\end{table}%

\subsection{Discussion}
Our \DMG{method} outperformed the \DMG{baselines} in target classification \DMG{mainly due to} two reasons. First, the classification does not depend on a clustering step. \DMG{This decreases the impact of cases shown in Fig.} \ref{fig:Clustering_errors} and \DMG{allows to handle objects that contain a single radar target (a common occurrence, especially for pedestrians, see Table} \ref{table:dateset}). 
Second, we included low-level radar data, which brings information of the speed distribution around the radar target. 
To demonstrate that this inclusion is beneficial, we showed that only using target-level data and only the third module of the network (\LLThigh) caused a significant drop in performance from \rev{0.70} to 0.61 average F1 score.
We examined the effect of removing absolute speed from the data too with \LLTnospeed. While the performance dropped, our network was still able to classify the radar targets by the relative speed distribution around them. 
The results of \LLThigh and \LLTnospeed proves that the relative velocity distribution (i.e. the low-level radar data) indeed contains valuable class information.
Interestingly, excluding RCS value did not have a significant impact on the performance. 
Based on our experiments, an ensemble of \rev{binary classifiers} results in less inter-class miss-classifications than using a single multi-class network. \looseness=-1

Note that even VRUs in occlusion (see Fig. \ref{subfig:sample_1}, \ref{subfig:sample_2}, \ref{subfig:sample_7}) are often classified correctly caused by the multi-path propagation of radar \cite{Palffy2019}. This, and its uniform performance in darkness/shadows/bright environments makes radar a useful complementary sensor for camera.
%
Typical errors are shown \DMG{in} Fig. \ref{fig:sample_errors}.
Radar is easily reflected by flat surfaces (e.g. side of cars) acting like mirrors, creating \textit{ghost targets}. E.g. \DMG{in} Fig. \ref{subfig:error_2} our ego-vehicle was reflected creating several false positives.
{Fig. \ref{subfig:error_5}} is an example of hard to categorize road users.
Many errors come from the confusion of \textit{car} and \textit{cyclist} caused by the similarity  of their Doppler signature and reflectivity, see Fig. \ref{subfig:error_6}. 
\DMG{Fig.} \ref{subfig:error_7} shows that a strong reflection nearby can mislead the classifier.
Since our method does not throw away single targets in a clustering step, it has to deal with more noise reflections than a cluster-wise method. However, the results in \textit{other} class suggest that it learned to ignore them. 

The combination of our network and the \DMG{clustering} step outperformed the baseline methods in the object detection task. This is mainly because by swapping the clustering and classifying steps, classes can be clustered with different parameters. That is a significant advantage of our pipeline, 
as instead of finding a single set of clustering parameters to handle each class, we can tune them separately to fit each, see Table \ref{tab:parameters}. This is especially useful in \textit{pedestrian} and \textit{car} classes, which are smaller/larger than the optimal spatial radius \rev{$ \spatialRadius = 1.2-1.3~m$} found for the baselines. However, this radius fits bicycles well, which results in good performance on the \textit{cyclists} class for \baselineSchumann both on target-level and object-level.
Fig. \ref{fig:baseline_vs_us} shows two examples. 
DBSCAN falsely separated the car and the bus into several clusters, but merged the pedestrians into a single one using the optimized parameters, which caused \baselineSchumann to fail.
Our method managed to classify each radar target individually and cluster them correctly (i.e. keep the vehicles in a single cluster, but separate the pedestrians) using the class-specific clustering parameters.
Although we used DBSCAN in this paper, \rev{we expect this advantage to stand using different types of clustering}. 
On Fig. \ref{subfig:cl_error_1} we show a single mis-classified radar target, probably reflected by the speed bump. 
The resulting false positive pedestrian detection is trade-off of setting $ \MinPoints $ to one for pedestrians. As mentioned, cyclists and cars are often confused. This is especially true if several cyclist ride side-by-side, see \ref{subfig:cl_error_1}, since their radar characteristics (extension, speed, reflectivity) are car-like. Both errors usually occur for a single frame only, and can be alleviated by a temporal filtering and tracking system. \looseness=-1

\begin{figure}
	\centering
	\vspace{-6pt}
	\includegraphics[width = 1\linewidth]{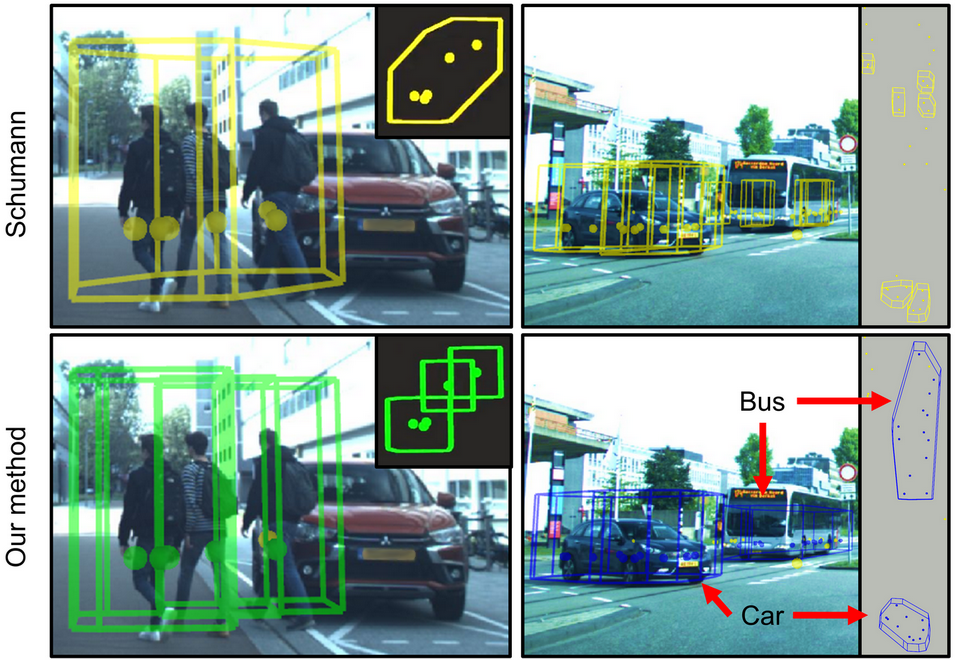}
	\caption{Challenging cases for clustering, camera and top view. DBSCAN falsely split the car and the bus but merged the pedestrians into a single cluster, making \baselineSchumann (top) fail. Our method (bottom) managed to classify the radar targets and cluster them correctly using class-specific parameters. 
	Yellow marks \textit{other} class.} 
	\label{fig:baseline_vs_us}
\end{figure}

\begin{figure}
	\begin{subfigure}{0.5\linewidth}
		\centering
		\adjincludegraphics[width=0.95\linewidth]{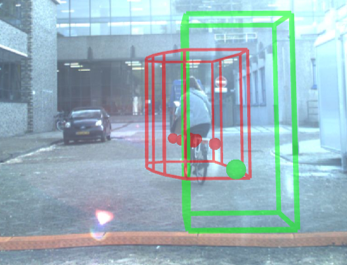}
		\caption{}
		\label{subfig:cl_error_1}
	\end{subfigure}%
	\begin{subfigure}{0.5\linewidth}
		\centering
		\adjincludegraphics[width=0.95\linewidth]{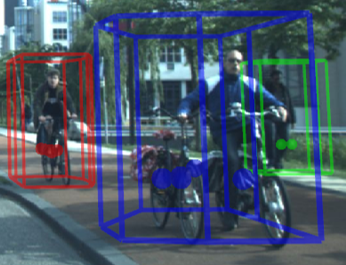}
		\caption{}
		\label{subfig:cl_error_2}
	\end{subfigure}%
	\caption{
		Examples of correct and incorrect object detections of our method.
		A mis-classified radar target triggered a false positive pedestrian detection on \subref{subfig:cl_error_1}. Bicycles moving side-by-side at the same speed 
		are detected as a car on \subref{subfig:cl_error_2}.
	} 
	\label{fig:cl_sample_errors}
\end{figure}
\section{CONCLUSIONS AND FUTURE WORK}
In this paper, we proposed a radar based, single-frame, multi-class road user detection method.
It exploits class information in low-level radar data by applying a specially designed neural network to a cropped block of the radar cube around each radar target and the target-level features. A \DMG{clustering step was introduced} to create object proposals. 

In extensive experiments on a real-life dataset we showed that the proposed method \DMG{improves upon} the baselines in target-wise classification by reaching an average F1 score of \rev{0.70} (vs. \rev{0.68} \baselineSchumann).
Furthermore, we demonstrated the importance of low-level features and ensembling in an ablation study.
We \DMG{showed that the proposed method outperforms the baselines overall in object-wise classification} by yielding an average F1 score of \rev{0.56} (vs. \rev{0.48} \baselineSchumann). 

Future work may include a more advanced \DMG{object clustering procedure}, e.g. by training a separate head of the network to  encode a distance metric for DBSCAN. Temporal integration and/or tracking of objects could further improve the method's performance and usability. \rev{Finally, extending the proposed framework to incorporate data from additional sensor modalities (e.g. camera, LiDAR) is worthwhile.}


\addtolength{\textheight}{-0.2cm}   



\section*{ACKNOWLEDGEMENT}
This work received support from the Dutch Science Foundation NWO-TTW, within the SafeVRU project (nr. 14667).
Andras Palffy {was also funded by the Tempus Public Foundation by means of a Hungarian Eotvos State Scholarship.}


\input{bib.def}
\bibliographystyle{IEEEtran}
\bibliography{radarbib}
\end{document}